\documentclass{article}
\pdfoutput=1

\PassOptionsToPackage{numbers}{natbib}
\usepackage[final]{nips_2017_arxiv}

\usepackage[utf8]{inputenc} 
\usepackage[T1]{fontenc}    
\usepackage{url}            
\usepackage{booktabs}       
\usepackage{amsfonts}       
\usepackage{nicefrac}       
\usepackage{microtype}      
\usepackage[toc,page]{appendix}

\usepackage{amsmath} 
\usepackage{amssymb}  
\usepackage{array}
\usepackage[usenames,dvipsnames,svgnames,table]{xcolor}
\usepackage{algorithm}
\usepackage{algpseudocode}
\algnewcommand\REQUIRE{\item[\textbf{Require:}]}%
\algnewcommand\INPUT{\item[\textbf{Input:}]}%
\algnewcommand\OUTPUT{\item[\textbf{Output:}]}%

\usepackage{wrapfig}

\usepackage[pdftex]{graphicx}
\graphicspath{{./images/}}
\DeclareGraphicsExtensions{.pdf,.jpg,.png}
\usepackage[caption=false,font=footnotesize]{subfig}

\usepackage{mathtools}

\DeclarePairedDelimiter\floor{\lfloor}{\rfloor}

\usepackage{amsthm}

\newtheorem{proposition}{Proposition}
\newtheorem{lemma}{Lemma}[section]

\newcommand*\diff{\mathop{}\!\mathrm{d}}

\newcommand{\E}{\mathbb{E}}

\newcommand{\bx}{\mathbf{x}}
\newcommand{\bz}{\mathbf{z}}
\newcommand{\Ex}{\E_{data}(D(\bx))}
\newcommand{\Exs}{\E_{data}(D^*(\bx))}
\newcommand{\Ez}{\E_{G}(D(\bx))}
\newcommand{\Ezs}{\E_{G}(D^*(\bx))}
\newcommand{\px}{p_{data}}
\newcommand{\pz}{p_{G}}

\usepackage{dsfont}

\usepackage{notoccite}
\usepackage{enumitem}
\setlist[itemize]{leftmargin=*,noitemsep, topsep=0pt}

\title{MAGAN: Margin Adaptation for Generative Adversarial Networks}

\author{
  Ruohan Wang \quad Antoine Cully\footnote[1]{} \quad Hyung Jin Chang\thanks{These authors contributed equally to this work} \quad Yiannis Demiris \\
  Personal Robotics Laboratory, Department of Electrical and Electronic Engineering\\ Imperial College London, United Kingdom\\
  \texttt{\{r.wang16, a.cully, hj.chang, y.demiris\}@imperial.ac.uk }\\
}

\newif\ifdraft\drafttrue
\ifdraft

\else

\fi

\begin{document}

\maketitle

\begin{abstract}
We propose the Margin Adaptation for Generative Adversarial Networks (MAGANs) algorithm, a novel training procedure for GANs to improve stability and performance by using an adaptive hinge loss function. We estimate the appropriate hinge loss margin with the expected energy of the target distribution, and derive principled criteria for when to update the margin. We prove that our method converges to its global optimum under certain assumptions. Evaluated on the task of unsupervised image generation, the proposed training procedure is simple yet robust on a diverse set of data, and achieves qualitative and quantitative improvements compared to the state-of-the-art.
\end{abstract}

\section{Introduction}
Generative Adversarial Networks (GANs)~\cite{goodfellow2014generative} are generative models known for their ability to sample from complex and intractable distributions, inherent in tasks such as realistic image generation from natural scenes. GANs are designed as a competitive game between the generator and discriminator network, whereby the generator tries to fool the discriminator with synthetic data, while the discriminator tries to differentiate the real data from the synthetic one. At its theoretical optimum, the generator produces samples indistinguishable from the real data. GANs have been applied to many interesting areas, including image super-resolution~\cite{16superres}, driving behavior modeling~\cite{kuefler17}, and data augmentation~\cite{shrivastava16}.

Training GANs remains a challenge. The quality of generated samples for complex distributions such as natural scenes is unsatisfactory, as they often contain visible artifacts and unrecognizable structures. In addition, GANs suffer from potential mode collapse whereby the generator only produces similar images~\cite{goodfellow17}, and a lack of convergence measures~\cite{arjovsky2017wasserstein}. Recent techniques including batch discrimination~\cite{salimans2016improved} and Wasserstein distance loss~\cite{arjovsky2017wasserstein} have addressed the above problems to varying degrees of success. However, they often introduce additional complexity such as weight clipping and new hyper-parameters, which demands significant tuning to achieve the best result.

Energy-based GANs (EBGANs)~\cite{zhao2016energy} use an auto-encoder as the discriminator, and define the energy of each sample as its reconstruction loss computed by the auto-encoder. The auto-encoder aims to assign lower energy to the real data than the synthetic one. EBGANs introduce a hinge loss objective function aimed at stabilizing the training, so that the discriminator could ignore synthetic samples with high energy. Finding an appropriate margin value is crucial for successful training and is dependent on both architecture choice and data complexity~\cite{zhao2016energy}.

In this paper, we propose Margin Adaptation for Generative Adversarial Networks (MAGANs), that automatically adapts the margin using the expected energy of the real data distribution, which improves the training stability by maintaining the equilibrium between the discriminator and the generator. 

The main contributions of this paper are:
\begin{itemize}
	\item A simple and robust training procedure for auto-encoder based GANs that adapts the hinge loss margin based on training statistics. This approach does not introduce any new hyper-parameters and removes the dependence on the margin hyper-parameter introduced by EBGANs.
    \item We prove that both EBGANs and MAGANs converge to their global optima under certain assumptions, and that MAGANs converge under more relaxed ones.
	\item A set of experiments that demonstrate the robustness and stability of our approach in producing visually appealing samples with relatively simple network architectures, and improvements over the state-of-the-art results on a diverse set of datasets.
\end{itemize}

\section{Related Work}
Generative Adversarial Networks (GANs)~\cite{goodfellow2014generative} are a class of generative sampling methods for modeling intractable distributions. Primary challenges of training GANs are summarized and discussed in~\cite{goodfellow17}. Among them, the difficulty in maintaining the equilibrium between the discriminator and generator often lead to training instability, as one network overpowers the other. Furthermore, visual inspection of generated samples is typically the only practical method to estimate convergence as the value of loss function fluctuates~\cite{goodfellow17, arjovsky2017wasserstein}. Various techniques have been proposed to improve GANs training~\cite{salimans2016improved,warde2017improving} to varying degrees of success. Notably, Wasserstein GANs (WGANs)~\cite{arjovsky2017wasserstein} provided the first convergence measure in GANs training using a loss function based on Wasserstein distance. To compute the Wasserstein distance, the discriminator uses weight clipping, which significantly limits the network capacity. Weight clipping has been later replaced with a gradient norm constraint in~\cite{gulrajani2017improved}. Unfortunately, these advances often introduce additional hyper-parameters, requiring significant tuning to achieve best results.

Auto-encoders~\cite{vincent2008extracting} are extensively used in GANs. In~\cite{warde2017improving}, the authors propose an auxiliary loss function to match latent features of real and synthetic samples, computed from an auto-encoder. In Plug-and-Play Generative Networks~\cite{nguyen16}, auto-encoders are used for computing the gradient of the log probability of real data to guide the sampling process. Energy-based GANs (EBGANs)~\cite{zhao2016energy} use an auto-encoder for the discriminator, which associates to each sample an energy value using the per-coordinate squared loss. Low energy samples are attributed to samples from the real data manifold, while high energy values are assigned to synthetic samples. EBGANs have been shown to minimize the total variation (TV) distance between the real and synthetic data distributions~\cite{arjovsky2017wasserstein}. EBGANs introduce a hinge loss objective function to stabilize the training by allowing the discriminator to ignore synthetic samples with energy larger than a predefined margin $m$ (see Eq. \ref{eq:loss}). Most recently, Boundary Equilibrium GANs (BEGANs)~\cite{began} extends EBGANs by proposing a loss function that aims to match the  energy of synthetic data to a fraction of the energy of real data. BEGANs share some similarities with the approach proposed in this paper, as they both monitor the expected energy of real and synthetic data to control training and generate visually pleasing and coherent samples. However, BEGANs formulate a compulsory trade-off between generation diversity and quality, and requires a more complex training procedure.

\section{Proposed Method}

\subsection{Definition}
Given a data sample $\bx \in \mathbb{R}^{N_{x}}$ of dimension $N_x$, a generated sample $G(\bz):\mathbb{R}^{N_{z}} \rightarrow \mathbb{R}^{N_{x}}$ and $\mathbf{z} \in \mathbb{R}^{N_{z}}$ of dimension $N_z$ generated from a known distribution, such as the normal distribution. Following~\cite{zhao2016energy}, we define the discriminator $D(\bx)$ as:
\begin{equation}
D(\bx) = || Dec(Enc(\bx)) - \bx ||
\end{equation}
where $D(\bx)$ is a deep auto-encoder function. The discriminator loss $L_D$ and the generator loss $L_G$ are defined as:
\begin{equation}
\label{eq:loss}
L_D(\bx, \bz)=D(\bx)+max(0, m-D(G(\bz))) \qquad \textrm{and} \qquad L_G(\bz)=D(G(\bz))
\end{equation}
where $m>0$ is a predefined hyper-parameter. 
We define $\Ex$ and $\Ez$, the expected energies of the real data distribution and of the synthetic data distribution respectively as:
\begin{equation}
\Ex=\int_{\bx}\px(\bx)\cdot D(\bx)\diff \bx \qquad \textrm{and} \qquad \Ez=\int_{\bx}\pz(\bx)\cdot D(\bx)\diff \bx
\end{equation}
where $\px, \pz$ denotes the probability that an arbitrary sample $\bx$ occurs in the real and synthetic data distributions respectively. In addition, we define two sets $S_1=\{\bx:\px(\bx)<\pz(\bx)\}$ and $S_2=\{\bx:\px(\bx)\geq \pz(\bx)\}$

\subsection{The MAGANs Model}
We have observed reliably in our experiments with EGBANs that the real and synthetic data distributions diverge (in terms of expected energy) during training, and that the quality of the generated samples stops improving visually (see Section~\ref{sec:exp:margin}). To overcome the divergence between the two distributions, we propose to adapt the margin $m$ to the expected energy of real data when we detect that the stalling of the generator loss. Intuitively, lowering the margin from $a$ to $b$ limits the power of the discriminator by preventing it from differentiating against synthetic samples with energy $b\leq D(G(\bz))\leq a$. This allows the generator to produce new adversarial examples without their energy being simultaneously raised by the discriminator. In turn, the discriminator is presented with new adversarial samples that better matches the expected energy of the real data.

We define when and how to adapt the margin in the two following section, and present the MAGANs training procedure in Algorithm \ref{alg:magan}.

{\begin{algorithm}\footnotesize
    \caption{MAGANs algorithm}
    \label{alg:magan}
    \begin{algorithmic}[1] 
    	\REQUIRE $\alpha$, the learning rate, $b$, the batch size, $N$ the training set size, $T_{max}$ the max number of training epochs, $w$, initial discriminator parameters, $\theta$, initial generator parameters
        	\State $m_0=0$
            \For{$t=1$ \textbf{to} 2}\Comment{Pre-train discriminator}
            	\For{$j=1$ \textbf{to} $\floor{N/b}$}
                  \State Sample $\{\mathbf{x}^{i}\}^{b}_{i=1}$ a batch from the real data
                  \State $g_w=\triangledown_w[\frac{1}{b}\sum_{i=1}^b D(\mathbf{x}^{i})]$\Comment{$\mathbf{z}^i$ ignored as $m_0=0$}
                  \State $w=w+\alpha\times Adamax(w, g_w)$
                \EndFor
            \EndFor
            \State $m_1=E(D(\mathbf{x})), S^{0}_{G}=\infty$
            \For{$t=1$ \textbf{to} $T_{max}$}
            	\State $S_{data}^t=0, S_{G}^t=0$\Comment{Collect statistics into $S$ to compute $E$}
            	\For{$j=1$ \textbf{to} $\floor{N/b}$}
                	\State Sample $\{\mathbf{x}^{i}\}^{b}_{i=1}$ a batch from the real data\Comment{Train the discriminator}
                	\State Sample $\{\mathbf{z}^{i}\}^{b}_{i=1}$ a batch from the prior samples
                	\State $g_w=\triangledown_w[\frac{1}{b}\sum_{i=1}^b (D(\mathbf{x}^{i})+max(0, m_t-D(G(\mathbf{z}^i))))]$
                	\State $S_{data}^t=S_{data}^t+\sum_{i=1}^b D(\mathbf{x}^{i})$ 
                	\State $w=w+\alpha\times Adamax(w, g_w)$
                	\State Sample $\{\mathbf{z}^{i}\}^{b}_{i=1}$ a batch from the prior samples\Comment{Train the generator}
                	\State $g_{\theta}=\triangledown_{\theta}[\frac{1}{b}\sum_{i=1}^b D(G(\mathbf{z}^i))]$
                	\State $\theta=\theta+\alpha\times Adamax(\theta, g_{\theta})$
                	\State $S_{G}^t=S_{G}^t+\sum_{i=1}^b D(G(\mathbf{z}^{i}))$
                \EndFor
                \If{$S_{data}^t/N<m_t$ \textbf{and} $S_{data}^t<S_{G}^t$ \textbf{and} $S^{t-1}_{G}\leq S_{G}^t$}\Comment{Update the margin}
                	\State $m_{t+1}=S^t_{data}/N$ 
                \EndIf
            \EndFor
    \end{algorithmic}
\end{algorithm}}

\subsubsection{When to Adjust the Margin}
We reduce the margin $m$ when the following three conditions hold:
\begin{equation}
E_{G}^{t-1}\leq E_{G}^{t} \qquad\text{and}\qquad E_{data}^{t}<m_t \qquad\text{and}\qquad E_{data}^{t}<E_{G}^{t}
\end{equation}
where $\E_G^t$ denotes $\Ez$ at the end of training epoch $t$. $\E_{data}^t$ is defined similarly for $\Ex$. The conditions capture the intuition that the discriminator should firstly attribute lower energy to the real data, which provides sufficient amount of samples with low energy that the generator could imitate from. To reduce the noise in computing the expected energy, we collect sample energies across each training epoch and compute the average at the end of each epoch.

\subsubsection{How to Set the Margin}
We choose $m_t=\E_{x}^t$ to guide the generator towards the real data distribution. In practice, the computation of expectation requires almost no additional resources as the expected energies of mini-batches are needed for the gradient computation. In order to get an initial estimate of the margin, we pre-train the discriminator with an auto-encoder objective for 2 epochs with only real samples (equivalent to setting $m=0$) and compute the expected energy of the real data.

The idea of using the expected energy to control training is also independently proposed in BEGANs~\cite{began}, where the loss function aims to match $E_{G}^t=\gamma E_{data}^t$ with $\gamma$ as a hyper-parameter meant to control the trade-off between diversity and quality. We observe that our method is different in several significant ways:
\begin{itemize}
	\item We show that our method converges to its global optimum where the real and synthetic data distributions match exactly. In contrast, BEGANs provide no such guarantee.
	\item Our method does not introduce new hyper-parameters. On the contrary, we remove the dependence on the margin hyper-parameter from EBGANs.
	\item Our method has a simpler training procedure, requiring no learning rate decay nor separate learning rates for the two networks.
\end{itemize}

\subsection{Theoretical Analysis of MAGAN}
\label{sec:magan_pr}
We show that both MAGANs and EBGANs converge to their global optima under certain assumptions, and that MAGANs converge under more relaxed conditions. We follow the same theoretical assumptions as in~\cite{goodfellow2014generative} for our analysis. Specifically, we assume that at each update step, 1) the discriminator is allowed to reach its optimum given the generator, 2) that $\pz$ is updated to improve its loss function, and that 3) $D$ and $G$ have infinite capacity. We also assume that $D^*(\bx)=0$ if $\px(\bx)=\pz(\bx)$, where $D^*$ denotes the optimal discriminator to a given generator. For the following demonstrations, we hypothesize that the expected energies can be computed at each iteration $t$.

\subsubsection{Convergence of EBGANs}
\label{sec:pr:ebgan}
\begin{lemma}
\label{pr:opti_d}
Let $D^*$ be the optimal discriminator with respect to a given generator, then $\Exs \leq \Ezs \leq m$. In particular, $\Exs=\Ezs$ if and only if $\px(\bx)=\pz(\bx)$ almost everywhere.
\end{lemma}
\vspace{-0.4cm}
\begin{proof}
Since $D^*$ is optimal with respect to $G$, 
\begin{equation}
\label{eq:opti_d}
 D^*(\mathbf{x})= 
\begin{dcases}
    m,& \text{if } \px(\mathbf{x})<\pz(\mathbf{x}) \\
    0,& \text{if } \px(\mathbf{x})>\pz(\mathbf{x})
\end{dcases}
\end{equation}
if $\px(\mathbf{x})=\pz(\mathbf{x})$, $D^*(\bx)$ is not unique and $0\leq D^*(\bx)\leq m$.
Equation \ref{eq:opti_d} is proven in~\cite[p.~11]{zhao2016energy}. Therefore, 
\begin{align}
\Exs - \Ezs &= \int_{S_1}(\px(\bx)-\pz(\bx))\cdot D^*(\bx)\diff \bx\\
&=m \cdot \int_{S_1}(\px(\bx)-\pz(\bx))\diff \bx \leq 0
\end{align}
\begin{equation}
\Ezs=\int_{\bx}\pz(\bx)D^*(\bx)\diff \bx \leq m\cdot \int_{\bx}\pz(\bx)\diff \bx=m
\end{equation} 
Lastly, if $\px(\bx)=\pz(\bx)\ \forall \bx$, it is clear that $\Exs=\Ezs$.
On the other hand, if $\Exs=\Ezs$, then
\begin{align}
\Exs - \Ezs &= m \cdot \int_{S_1}(\px(\bx)-\pz(\bx))\diff \bx=0
\end{align}
Since $\px(\bx)-\pz(\bx)$ is negative for $\bx \in S_1$, the equality only holds if $\int_{S_1}\diff \bx=0$. This implies that $\px(\bx)=\pz(\bx)$ almost everywhere\footnote[1]{The last step is proven in Lemma 2 of~\cite[p.~11]{zhao2016energy}}.
\end{proof}

\begin{lemma}
\label{pr:invar_g}
Let $G_t$ and $D^*_t$ be the generator and its optimal discriminator at update step $t$ of EBGANs training, then $\E_{G_{t+1}}(D^*_t(\bx))=\E_{G_{t+1}}(D^*_{t+1}(\bx))$,
\end{lemma}
\vspace{-0.4cm}
\begin{proof}
By assumption that $D^*(\bx)=0$ \text{ if } $\px(\bx)=\pz(\bx)$, Equation \ref{eq:opti_d} reduces to 
\begin{equation}\label{eq:representation}
 D^*(\bx)= 
\begin{dcases}
    m,& \text{if } \px(\bx)<\pz(\bx) \\
    0,& \text{if } \px(\bx) \geq \pz(\bx) \\
\end{dcases}
\end{equation}
As $D^*_t$ and $D^*_{t+1}$ do not change $S_1$ or $S_2$ (only updates to G change $S_1$ and $S_2$), 
\begin{equation*}
\E_{G_{t+1}}(D^*_t(\bx))=0\cdot \pz(S_2)+m \cdot \pz(S_1)=\E_{G_{t+1}}(D^*_{t+1}(\bx))
\end{equation*}
\end{proof}

\begin{proposition}
\label{pr:ebgan}
If at each update step, the optimal discriminator is reached given the generator, and $\pz$ is updated to reduce $\Ez$, EBGANs converge to $\px(\bx)=\pz(\bx)$ almost everywhere.
\end{proposition}
\vspace{-0.4cm}
\begin{proof}
The two conditions combined translates to $\E_{G_{t+1}}(D^*_t(\bx))<\E_{G_{t}}(D^*_t(\bx))\ \forall t$. By Lemma \ref{pr:invar_g}, $\E_{G_{t+1}}(D^*_t(\bx))=\E_{G_{t+1}}(D^*_{t+1}(\bx))$.
Therefore,
\begin{equation}
\E_{G_{t+1}}(D^*_{t+1}(\bx))=\E_{G_{t+1}}(D^*_t(\bx)) < \E_{G_{t}}(D^*_t(\bx))
\end{equation}
This implies that the expected energy of synthetic data strictly decreases at each update step.

From~\cite{arjovsky2017wasserstein}, we rewrite the loss function of EBGANs in minimax form under optimal discriminators,
\begin{equation}
\min_{\pz} \max_{0\leq D(\bx)\leq m} \Ez-\Ex
\end{equation}
Noting that the following expression is convex with respect to $\pz$\footnote[2]{Derivation in the Supplementary Material.}
\begin{equation}
\Ezs-\Exs=\frac{m}{2}\int_\bx|\pz(\bx)-\px(\bx)|\diff\bx
\end{equation}

and that there exists a unique global optimum when the two distributions exactly match~\cite{zhao2016energy,arjovsky2017wasserstein}.  Therefore, with sufficiently small update steps of $\pz$, $\pz$ converges to the global optimum, which implies $\px(\bx)=\pz(\bx)$ almost everywhere.
\end{proof}

\subsubsection{Convergence of MAGANs}
\begin{lemma}
\label{pr:margin}
In MAGAN, $m$ strictly decreases and m converges to 0 as long $\px$ and $\pz$ do not match exactly.
\end{lemma}
\vspace{-0.4cm}
\begin{proof}
When $m_t$ is updated to $\E_{data}(D^*_t(\bx))$, we have  $m_t=\E_{data}(D^*_t(\bx))=m_{t-1}\cdot \px(S_1)$ (according to equation~\ref{eq:representation}). Therefore, $m$ strictly decreases and converges to $0$, as $\px(S_1)<1$.
\end{proof}
\begin{proposition}
\label{pr:magan}
If at each update step, the optimal discriminator is reached given the generator, and $\pz$ is updated along the direction of $-\frac{\partial \Ezs}{\partial \pz(\bx)}$, MAGANs converge to $\px(\bx)=\pz(\bx)$ almost everywhere.
\end{proposition}
\vspace{-0.4cm}
\begin{proof}
We first note that updating $\pz$ along the direction of $-\frac{\partial \Ezs}{\partial \pz(\bx)}$ corresponds to a gradient descent step of G towards reducing the generator loss function. We do not assume that the update steps reduce $\Ez$ at each step, but only that the directions of the update steps is appropriate with respect to $\pz$.





We split the proof into two cases: 1) As long as the update steps are small enough such that $\Ez$ is reduced after each step, the margin remains unchanged and MAGANs converge to its global optimum (similarly to Proposition~\ref{pr:ebgan}).

2) If $\Ez$ increases after a step $t$ because of a too large step, the margin update condition $\E_{G_{t+1}}(D^*_t(\bx))\geq \E_{G_{t}}(D^*_{t}(\bx))$ is met, and the other two conditions are satisfied by Lemma \ref{pr:opti_d}).

Using the Radon–Nikodym Theorem, we can show that the magnitude of the update steps is proportional to $m_t$: 
\begin{equation}
 \frac{\partial \E_{G_{t}}(D^*_t(\bx))}{\partial \pz(\bx)}= 
\begin{dcases}
    m_t,& \text{if } \bx \in S_1 \\
    0,& \text{if } \bx \in S_2 \\
\end{dcases}
\end{equation}

This implies that the margin adaptation reduces the magnitude of the update steps, and as $m$ strictly decreases and converges to $m_\infty\rightarrow 0$ (according to Lemma \ref{pr:margin}), therefore it is guaranteed to find a $m$ small enough to meet the conditions of the first case above, specifically that the update steps are sufficiently small to reduce $\Ez$ at each step.
\end{proof}


While the effects of margin adaptation share some connections with learning rate decay from a theoretical standpoint, its empirical effects are subtler, as the adjustments are based on the on training statistics. In particular, we hypothesize that margin adaptation maintains the equilibrium between the generator and the discriminator, which improves training stability and consequently the generator performance. Lowering the margin temporarily limits the power of the discriminator by removing more synthetic samples from it loss function, until the generator catches up by generating more consistently samples with lower energies. At the same time, the margin reduction also protects the discriminator from wasting model capacity on synthetic samples likely of low quality (i.e. of high energy) and thus providing better gradients for the generator.

\section{Experimental Results}
\label{sec:exp}
For all experiments in this paper, we use a deep convolutional generator, analogous to DCGAN's~\cite{dcgan}. For the discriminators, we use a fully-connected auto-encoder for the MNIST dataset, and a fully convolutional one forthe  CIFAR-10 and the CelebA datasets. The convolutional auto-encoder is composed of strided convolution for encoding and fractional-strided convolutions for decoding. All models are trained with Adamax~\cite{adam} with a fixed learning rate of 0.0005, and momentum $\beta_1$ of 0.5, and a batch size of 64. Exact model architectures are reported separately for each dataset in the Supplementary Materials.
The method requires no other techniques such as batch normalization~\cite{batchnorm}, or layer-wise noises~\cite{zhao2016energy} to help with the training. We sample $z$ from $\mathcal{N}(0, 1)$ and determine $N_{z}$ such that the number of parameters in the discriminator and generator are roughly equal. The code used for the experiments is available on GitHub\footnote[3]{\url{https://github.com/RuohanW/magan}}.

\subsection{Quantitative Evaluation}
\begin{figure*}
\centering
\subfloat[]{\includegraphics[width=2in]{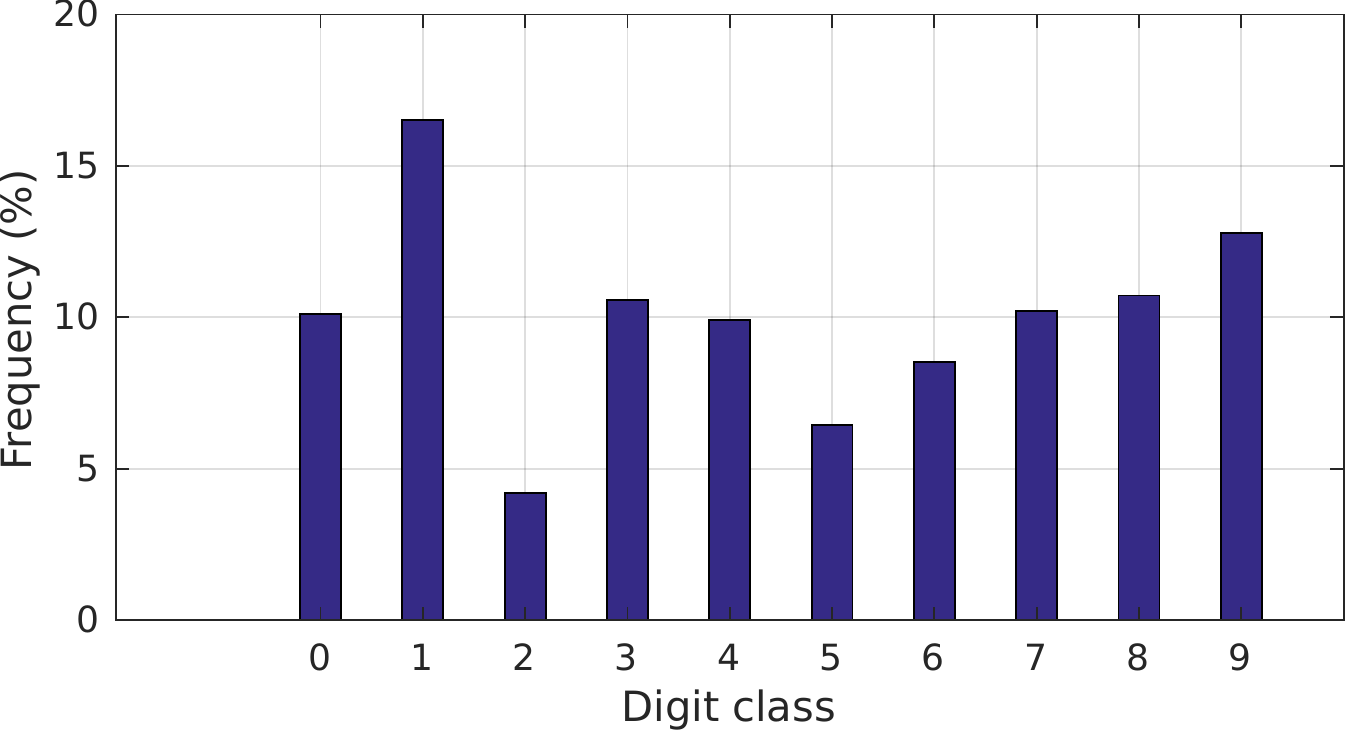}\label{fig:mnist_freq}}
\hfill
\subfloat[]{\includegraphics[width=1.35in]{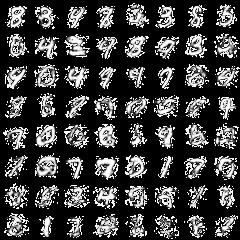}\label{fig:mnist_eb}}
\hfill
\subfloat[]{\includegraphics[width=1.35in]{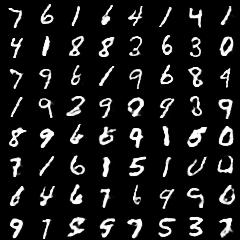}\label{fig:mnist}}
\hfill
\caption{(a) MNIST generated samples distribution. The distribution shows relatively balanced generation among all 10 classes of the MNIST. (b) Noisy random generation using a fixed margin (equivalent to EBGANs). (c) Random generation from our proposed method. (b) and (c) differ only by the proposed adaptive margin.}
\label{fig:mnist_all}
\end{figure*}

We first show that our method generates diverse samples with no mode collapse. In the first experiment, our architecture is trained on the MNIST dataset. A random mini-batch of samples is shown in Figure \ref{fig:mnist}. To show that the model does not suffer from mode collapse, we trained a three-layer convolutional classifier on the MNIST separately (98.7\% accuracy on MNIST test set) and used the classifier on 50000 samples generated with our architecture. The results are shown in Figure \ref{fig:mnist_freq}.

The results show that each of the 10 classes of MNIST is present, while 5 out of 10 classes are generated with a probability close to 10\%, matching the original data distribution. The classes of "ones" and "twos" are respectively the most over and under-represented with 16.5\% and 4.2\% of the distribution, which may be explained by the similarity between the "twos" and other classes, and the simplicity of ones.

To quantify the sample diversity and quality of our results, we compute the inception score~\cite{salimans2016improved} on the MNIST samples generated by our architecture. The inception score is a heuristic commonly used with GANs to measure single sample quality and diversity using a standard pre-trained inception network. We follow previous works to compute the score with 10 batches of 5000 independent generated samples, using the evaluation script from~\cite{salimans2016improved}. We emphasize that in order to compute the inception score for the MNIST, we replaced the pre-trained inception model with the aforementioned convolutional MNIST classifier. To compare against EBGANs, we used code from~\cite{ebgan_tf}, a EBGANs implementation specifically tuned for MNIST generation. We report that our model achieves a score of $7.52\pm 0.03$ against $7.14\pm 0.04$ from EBGANs. The real samples from the MNIST test set achieves a score of $9.09\pm 0.08$.

\begin{wraptable}{r}{6.0cm}
  \caption{CIFAR-10 Inception score comparison}
  \label{tab:cifar}
  \centering
  \resizebox{1.0\linewidth}{!}{	
  \begin{tabular}{lll}
    \toprule
   	Method     & Score $\pm$ Std\\
    \midrule
    Real data & $11.24\pm 0.12$\\
	DFM~\cite{warde2017improving} & $7.72\pm0.13$\\
    EGANs~\cite{dai2017calibrating} & $7.07\pm 0.10$\\
	BEGANs~\cite{began} & 5.62\\
	ALI~\cite{ali} (from~\cite{warde2017improving}) & $5.34\pm 0.05$\\
	Improved GANs~\cite{salimans2016improved} & $4.36\pm 0.04$\\
    MIX + WGANs~\cite{mix_wgan} & $4.04\pm 0.07$\\
    Wasserstein GANs~\cite{arjovsky2017wasserstein} (from~\cite{mix_wgan}) & $3.82\pm 0.06$\\
	\textbf{MAGANs (proposed method)} & \textbf{6.40 $\pm$ 0.03}\\
    \bottomrule
  \end{tabular}  }
\end{wraptable} 

To further quantify the performance of our proposed architecture, we trained our model on the CIFAR-10 dataset and computed the inception score.  The result is presented in Table \ref{tab:cifar}, and compared against other state-of-the-art unsupervised GANs. For fair comparison, we do not consider models using labeled data.

With the exception of Denoising Feature Matching (DFM)~\cite{warde2017improving} (DFM) and EGAN~\cite{dai2017calibrating}, our method outperforms all other methods by a large margin. Both DFM and EGANs introduces auxiliary training objectives to improve the performance of the generator, which are beyond the scope of this paper. Both works are compatible with our framework and represent possible directions for future investigations. Preliminary tests have shown promises in this direction.

\subsection{Qualitative Evaluation}
\label{sec:exp_quant}
For qualitative comparisons, we use the CelebA dataset, as generating realistic, highly detailed and flawless faces of different poses remains an open challenge in GANs. We compare the generation results against those of BEGANs and EBGANs, as they are both state-of-the-art auto-encoder GANs, directly comparable to our method. For EBGANs, we directly use the results from~\cite{zhao2016energy}. For BEGANs, we use the code from~\cite{began_tf}, which is specifically tuned for the CelebA dataset, as the original paper does not release source code and used a private dataset. We also standardized the cropping of training images for all the compared architectures.

\begin{figure*}
\centering
\subfloat[]{\includegraphics[width=1.8in]{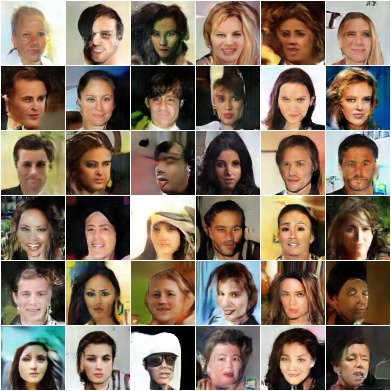}}
\hfill
\subfloat[]{\includegraphics[width=1.8in]{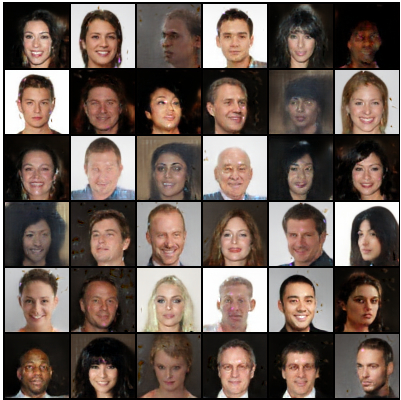}}
\hfill
\subfloat[]{\includegraphics[width=1.8in]{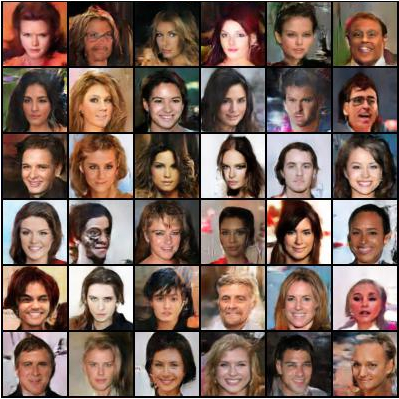}}
\hfill
\caption{(a) EBGANs CelebA generation taken from~\cite{zhao2016energy}. (b) BEGANs CelebA generation based on~\cite{began_tf}. (c) CelebA generation from our method. Results from BEGANs and our method are from a random mini-batch of generates samples respectively. Best viewed in color and enlarged. More samples are available in the Supplementary Material.}
\label{fig:celebA}
\end{figure*}

Compared against EBGANs, our results exhibit more detailed and coherent faces, and appear to better capture the symmetric property of faces and their appearances under different poses. Both EBGANs and our results show rich and diverse backgrounds, consistent with backgrounds contained in the training set. We follow closely to the parameter and architecture choices of EBGANs to attribute the improvements in visual quality to the proposed margin adaptation.

Our method generates very different samples compared to BEGANs. We first note that BEGANs results are comparable to those presented in the original paper, with similar characteristics. BEGANs mostly generate uniform background between black and white, compare to the rich variety of backgrounds from the training data, and from our results. While both BEGANs and MAGANs generate visually appealing faces,  our results demonstrate more diversity in terms of shape, hairstyle and color.

\subsection{The Effects of Margin Adaptation}
\label{sec:exp:margin}
\begin{figure*}
\centering
\subfloat[Comparison of real samples energy between proposed method and EBGAN]{\includegraphics[width=2.6in]{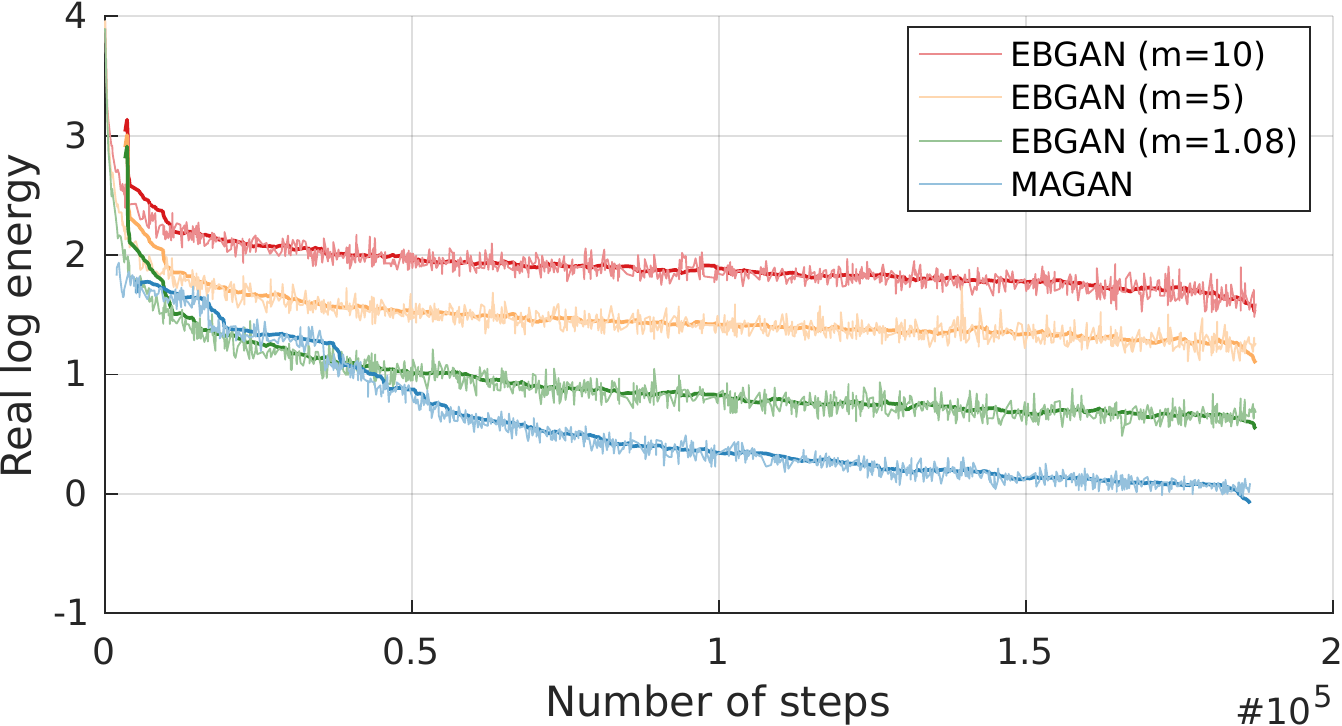}}
\hfil
\subfloat[Comparison of synthetic samples energy between proposed method and EBGAN]{\includegraphics[width=2.6in]{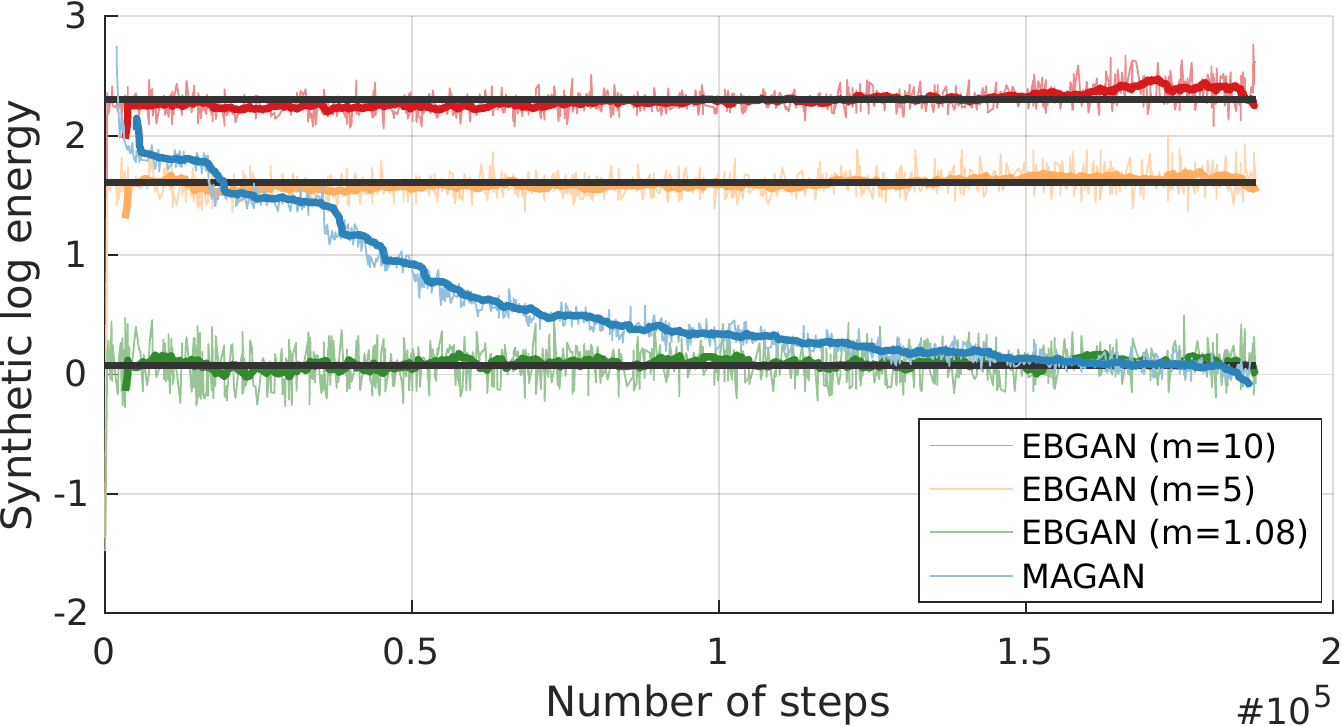}}
\hfil
\caption{EBGANs have diverging energy for real and synthetic samples on the MNIST dataset. The black lines in (b) denotes EBGANs margins. The generator loss rises gradually and steadily above the preset margin. Real and synthetic data energy decreases in tandem in MAGANs. Best viewed in color.}
\label{fig:loss_cmp}
\end{figure*}

To further verify the effects of margin adaptation, we compared our method against EBGANs on the MNIST dataset, while keeping all other parameters and network architectures identical. We trained three EBGANs models by setting the margin at 10, 5 and 1.08 (the final margin value obtained by MAGANs) respectively. Figure \ref{fig:loss_cmp} shows the evolutions of real and synthetic sample energy during training, while Figure \ref{fig:mnist_all} shows the comparison of random samples generated from the two methods. All three EBGANs models produce samples of similar quality, shown in Figure \ref{fig:mnist_eb}.

Qualitatively, the proposed method generates crisp and sharp samples, compared against the noisy samples produced by EBGANs. Quantitatively, real and synthetic sample energy decreases in tandem for the proposed method. In contrast, the energies of real and synthetic samples diverge in EBGANs. In particular, the energy of synthetic samples, equivalent to the generator loss, does not decreases over time, which causes difficulty in using the loss metric as an estimate of the training progress. Similar observations have been observed for standard GANs, whereby the Jensen-Shannon distance between the two distributions, a measure of similarity between two distributions, also worsen as the training proceeds~\cite{arjovsky2017wasserstein}. Along with the qualitative and quantitative improvements demonstrated in the previous sections, the results strongly suggest that margin adaptation improves both the stability and quality of GANs training.

\section{Conclusion and Future Work}
We have presented Margin Adaptation for GANs, a novel training procedure for auto-encoder GANs. We have shown theoretically that margin adaptation allows GANs to converge under more general conditions, which improves training stability and both qualitative and quantitative performance compared to the state-of-the-art. For future work, we wish to explore the use of margin adaptation in other GANs frameworks to test its general applicability.

\bibliographystyle{unsrt}
\bibliography{magan}

\begin{appendices}
\section{Technical Details for Section \ref{sec:pr:ebgan}}
We show that $\Ezs-\Exs=\frac{m}{2}\int_\bx|\pz(\bx)-\px(\bx)|\diff\bx$.
\begin{proof}
we first show that $\int_{S_1} (\pz(\bx)-\px(\bx))\diff\bx=\frac{1}{2}\int_\bx|\pz(\bx)-\px(\bx)|\diff\bx$
\begin{align*}
\int_{S_1} (\pz(\bx)-\px(\bx))\diff\bx &+\int_{S_2} (\pz(\bx)-\px(\bx))\diff\bx=0\\
\int_{S_1} (\pz(\bx)-\px(\bx))\diff\bx &=\int_{S_2} (\px(\bx)-\pz(\bx))\diff\bx\\
\int_{S_1} |\pz(\bx)-\px(\bx)|\diff\bx &=\int_{S_2} |\pz(\bx)-\px(\bx)|\diff\bx\\
\int_{S_1} |\pz(\bx)-\px(\bx)|\diff\bx &=\frac{1}{2}(\int_{S_1} |\pz(\bx)-\px(\bx)|\diff\bx+\int_{S_2} |\pz(\bx)-\px(\bx)|\diff\bx)\\
\int_{S_1} (\pz(\bx)-\px(\bx))\diff\bx &=\frac{1}{2}\int_\bx|\pz(\bx)-\px(\bx)|\diff\bx
\end{align*}
Therefore,
\begin{align*}
\Ezs-\Exs &=m\int_{S_1} (\pz(\bx)-\px(\bx))\diff\bx\\
&=\frac{m}{2}\int_\bx|\pz(\bx)-\px(\bx)|\diff\bx
\end{align*}
\end{proof}

\section{Supplementary Details for Experimental Setups}
We use FC, CV and DC to denote fully-connected, convolution and deconvolution layers. For example, FC(128) denotes a fully-connected layer with a 128-unit output. CV(64, 4c2s) denotes a convolution layer with 64 output feature maps, using 4x4 kernels with stride 2. DC are defined similarly as CV.

\subsection{MNIST Experiment}
Discriminator: Input-FC(256)-FC(256)-FC(784)

Generator: Input-DC(128, 7c1s)-DC(64, 4c2s)-DC(1, 4c2s)

We set $N_z=50$, to roughly balance the network capacity of the generator and the discriminator. All internal activations use RELU units while the output layers of both the discriminator and the generator use sigmoid units. The same architecture is used for Section \ref{sec:exp:margin}.

\subsection{CelebA Experiment}
Discriminator: Input-CV(64, 4c2s)-CV(128, 4c2s)-CV(256, 4c2s)-CV(512, 4c2s)-DC(256, 4c2s)-DC(128, 4c2s)-DC(64, 4c2s)-DC(3, 4c2s)

Generator: Input-DC(512, 4c1s)-DC(256, 4c2s)-DC(128, 4c2s)-DC(64, 4c2s)-DC(3, 4c2s)

We set $N_z=350$ to approximately balance the capacity of the two networks. The discriminator uses Leakly RELU for internal activations, while the generator uses RELU. The output layers of both networks use tanh units.

\subsubsection{CIFAR-10 Experiment}
Discriminator: Input-CV(64, 3c1s)-CV(128, 3c2s)-CV(128, 3c1s)-CV(256, 3c2s)-CV(256, 3c1s)-CV(512, 3c2s)-DC(256, 3c2s)-DC(256, 3c1s)-DC(128, 3c2s)-DC(128, 3c1s)-DC(64, 3c2s)-DC(3, 3c1s)

Generator: Input-DC(512, 4c1s)-DC(256, 3c2s)-DC(256, 3c1s)-DC(128, 3c2s)-DC(128, 3c1s)-DC(64, 3c2s)-DC(3, 3c1s)

We set $N_z=320$. The discriminator uses Leakly RELU for internal activations, while the generator uses RELU. The output layers of both networks use tanh units.

\section{Additional Generated Samples For CIFAR-10 and CelebA}
All samples shown below are random mini-batches of size 64.
\begin{figure*}
\centering
\subfloat[]{\includegraphics[width=0.7\textwidth]{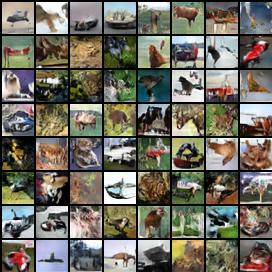}}
\hfil
\subfloat[]{\includegraphics[width=0.7\textwidth]{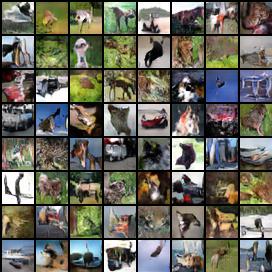}}
\hfil
\caption{CIFAR-10 samples generated by MAGANs}
\end{figure*}

\begin{figure*}
\centering
\subfloat[]{\includegraphics[width=0.7\textwidth]{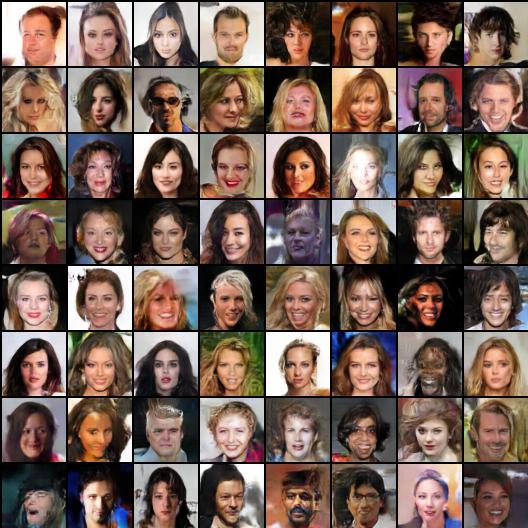}}
\hfil
\subfloat[]{\includegraphics[width=0.7\textwidth]{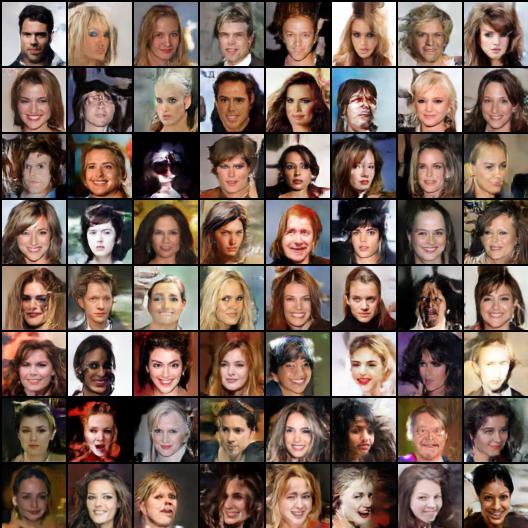}}
\hfil
\caption{CelebA samples generated by MAGANs}
\end{figure*}

\begin{figure*}
\centering
\subfloat[]{\includegraphics[width=0.7\textwidth]{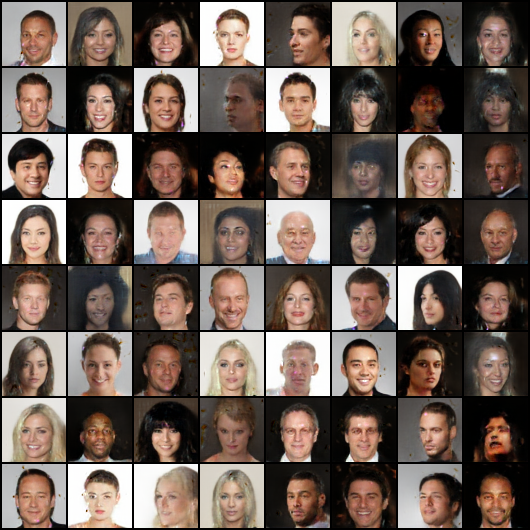}}
\hfil
\subfloat[]{\includegraphics[width=0.7\textwidth]{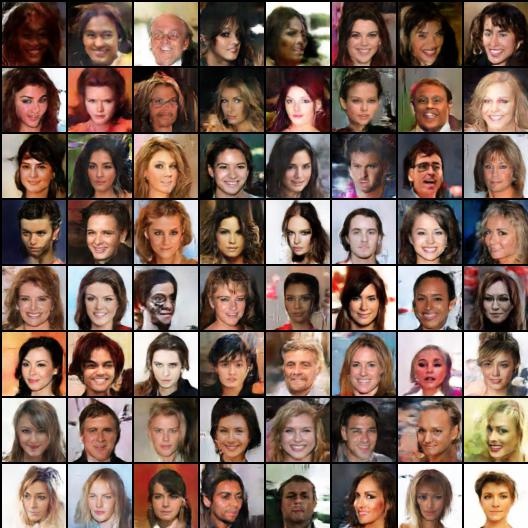}}
\hfil
\caption{More CelebA Comparison Against BEGANs}
\end{figure*}
\end{appendices}

\end{document}